\pdfoutput=1

\documentclass[11pt]{article}
\usepackage{authblk}
\usepackage{array}
\usepackage{acl}
\usepackage{times}
\usepackage{latexsym}
\usepackage[T1]{fontenc}
\usepackage[utf8]{inputenc}
\usepackage{microtype}

\usepackage{amsmath}
\usepackage{xcolor}
\usepackage{upquote}
\DeclareMathOperator*{\argmin}{arg\,min}
\usepackage{arydshln}
\usepackage{bibentry}
\usepackage{times}
\usepackage{latexsym}
\usepackage{booktabs}
\usepackage{textcomp}
\usepackage{graphicx}
\usepackage{multirow} 
\usepackage{enumitem}
\usepackage{tkz-base}
\usepackage{booktabs}
\usepackage{rotating}

\newcommand{\bart}{\textsc{Bart}}
\newcommand{\tldr}{\textsc{Tldr}}
\newcommand{\rgo}{\textsc{Rouge-1}}
\newcommand{\rgt}{\textsc{Rouge-2}}
\newcommand{\rgl}{\textsc{Rouge-L}}
\newcommand{\loss}{\ell}
\newcommand{\losslrg}{\mathcal{L}}

\newcommand{\currsum}{\textsc{CurrSum}}
\newcommand{\sentsum}{\textsc{SentSum}}
\newcommand{\currsentsum}{\textsc{CurrSentSum}}
\newcommand{\superl}{\textsc{SuperLoss}}
\usepackage{inconsolata}

\title{Curriculum-Guided Abstractive Summarization}

\makeatletter
\def\thanks#1{\protected@xdef\@thanks{\@thanks
        \protect\footnotetext{#1}}}
\makeatother
\author[$\heartsuit^{\ast}$]{\textbf{Sajad Sotudeh} \thanks{* Work partially done during the internship at Adobe Research.}}

\author[$\spadesuit$]{\textbf{Hanieh Deilamsalehy}}
\author[$\spadesuit$]{\textbf{Franck Dernoncourt}}

\author[$\heartsuit$]{\textbf{Nazli Goharian}}

\affil[$\heartsuit$]{IRLab, Georgetown University}
\affil[ ]{\texttt{\{sajad,nazli\}@ir.cs.georgetown.edu}}

\affil[$\spadesuit$]{Adobe Research}
\affil[ ]{\texttt{\{deilamsa,franck.dernoncourt\}@adobe.com}}

\begin{document}
\maketitle
\begin{abstract}
Recent Transformer-based summarization models have provided a promising approach to abstractive summarization. They go beyond sentence selection and extractive strategies to deal with more complicated tasks such as novel word generation and sentence paraphrasing. Nonetheless, these models have two shortcomings: (1) they often perform poorly in content selection, and (2) their training strategy is not quite efficient, which restricts model performance. In this paper, we explore two orthogonal ways to compensate for these pitfalls. First, we augment the Transformer network with a sentence cross-attention module in the decoder, encouraging more abstraction of salient content. Second, we include a curriculum learning approach to reweight the training samples, bringing about an efficient learning procedure. Our second approach to enhance the training strategy of Transformers networks makes stronger gains as compared to the first approach. We apply our model on \textit{extreme} summarization dataset of \textit{Reddit TIFU} posts. We further look into three cross-domain summarization datasets (\textit{Webis-TLDR-17}, \textit{CNN/DM}, and \textit{XSum}), measuring the efficacy of curriculum learning when applied in summarization. Moreover, a human evaluation is conducted to show the efficacy of the proposed method in terms of qualitative criteria, namely, fluency, informativeness, and overall quality. 
\end{abstract}

\section{Introduction}

Text summarization systems aim to condense a piece of text to a shorter form that preserves the major information within the original text. This task is broadly done in two ways: (1) extractive~\cite{Nallapati2017SummaRuNNerAR, Xiao2019ExtractiveSO, Xu2020DiscourseAwareNE} which assembles the salient sentences directly from the source text, and (2) abstractive~\cite{elikyilmaz2018DeepCA, Lebanoff2018AdaptingTN, Liu2019TextSW, Zou2020PretrainingFA, sotudeh-etal-2021-tldr9} that involves paraphrasing, and generating novel words that are not present in the source text. 
Recent efforts have also tackled the task using hybrid  models~\cite{See2017GetTT, Hsu2018AUM, Chen2019MultiTaskLF, MacAvaney2019SIG}. 

Over the last decade, neural summarization models based on RNN~\cite{Hochreiter1997LongSM}, and Transformers~\cite{Vaswani2017Att} have achieved promising results on text summarization. The recent success of pre-trained language models on a wide variety of downstream tasks, including summarization, has driven the current state-of-the-art to a new level. While such approaches generate fluent summaries, 
 a few studies have recognized \textit{content selection} as their pitfall~\cite{Gehrmann2018BottomUpAS,Narayan2020StepwiseES}, restricting the model performance at generating \textit{informative} summaries. In this research, we aim to extend the decoder by inducing sentential information as a \textit{saliency signal} that can come in handy when summarizing the source. 

Large-scale deep neural models are often hard to train; leaning on intricate heuristic set-ups which can be time-consuming and expensive to tune~\cite{Gong2019EfficientTO, Chen2021EarlyBERTEB}. This is specially the case for the Transformers which have been shown to consistently outperform the RNN networks when rigorously tuned~\cite{Popel2018TrainingTF}, but also require heuristics such as specialized learning rates and large-batch training~\cite{Platanios2019CompetencebasedCL}. 
In this paper, we attempt to overcome the mentioned problem by introducing a \textit{curriculum learning (CL)} strategy for training the summarization model, leading to improved convergence time, and performance. Inspired by humans' teaching style, \textit{curriculum learning} suggests to move the teaching process from easier samples to more difficult ones, and dates back to the nineties~\cite{Elman1993LearningAD}. The driving idea behind this approach is that networks can accomplish better task learning when the training instances are exposed to the network in a specific order, from easier samples to more difficult ones~\cite{Chang2021DoesTO}. In the context of neural networks, this process can be thought as a technique that makes the network robust to getting stuck at local optima, which is more likely in the early stages of training process. 
Systems equipped with curriculum learning have been reported to show strong generalization, faster convergence time, and even improved model performance~\cite{Platanios2019CompetencebasedCL}.

After identifying the two drawbacks above (i.e., content selection and inefficient training) of Transformer-based networks, we develop our summarization framework to address these shortcomings. To remedy the first drawback, we propose to augment the transformer decoder with a \textit{sentence cross-attention layer}, encouraging the decoder to pay more attention to salient sentences of the source text while generating the abstractive summary. For the second, we supply the summarization model with curriculum learning objectives. We, specifically, utilize the \superl~\cite{Castells2020SuperLossAG} function that falls into the family of confidence-aware curriculum learning techniques, introducing a new parameter called confidence (i.e., $\sigma$) to the network. While learning this parameter is inefficient, especially on the abundance of training instances such as in summarization, \superl{} bridges this limitation by directly using the converged value of confidence at a specific learning state. We validate our model on the \textit{extreme summarization} task~\cite{Narayan2018DontGM}, where the aim is to produce a one-sentence summary in extreme compression and high abstraction. To this end, we make use of \textit{Reddit TIFU}~\cite{Kim2019AbstractiveSO} and Webis-TLDR-17~\cite{volske-etal-2017-tl} datasets containing 42k, and 4M instances, respectively, with each pair including a Reddit post along with its \tldr\footnote{\tldr{} is the abbreviation of ``Too Long, Didn't Read''.} summary. To measure our model's cross-domain performance, we further report model performance on CNN/DM~\cite{See2017GetTT} and XSum~\cite{Narayan2018DontGM} large-scale news datasets. We show that the inclusion of curriculum learning allows for a remarkable performance of neural Transformer-based summarizers. We further carry out a comprehensive human evaluation to examine the efficacy of our model in terms of three qualitative metrics: fluency, informativeness, and overall quality. 

\section{Related Work}
\textbf{Pre-trained Language Modeling.}
During the last few years, self-supervised pre-trained language models have gained increased attention from research community due to their considerable improvements in a variety of NLP tasks. Different variants of such models are pre-trained on a large amount of unlabeled data~\cite{Devlin2019BERTPO, Liu2019RoBERTaAR, Peters2018DeepCW}, each with various pre-training objectives. While such models are inherently proposed to perform language modeling task, it has been made possible to fine-tune them on a wide range of downstream NLP tasks, summarization being one of them. \citet{Liu2019TextSW} were the first to fine-tune \textsc{Bert} for summarization task. They specifically proposed three variants of \textsc{BertSum} including \textsc{BertSumExt} for extractive summarization, \textsc{BertSumAbs} for abstractive summarization, and \textsc{BertSumExtAbs} which is a two-stage fine-tuning approach, exploiting extractive and abstractive objectives. Following up this line of research, ~\citet{zhang2019pegasus} proposed \textsc{Pegasus} with pre-training objectives specific for text summarization and achieved state-of-the-art results on 12 downstream summarization tasks. In a parallel line, \citet{Lewis2020BARTDS} proposed \textsc{Bart} and showed its efficacy on language generation tasks such as text summarization. Unlike \textsc{BertSum} that uses merely pre-trained \textsc{Bert} encoder, \textsc{Pegasus} and \textsc{Bart} exploit both pre-trained encoder and decoder for language generation. 

\noindent \textbf{Sentence-guided Summarization.}
Using sentence representations as extractive signals along with token embeddings in neural sequence-to-sequence models has a recent history. This idea is inspired by the fact that while the general encoder-decoder frameworks produce fluent targets, they often fall short in content selection~\cite{Sotudeh2020AttendTM}. A few works have noted that this problem can be addressed by combining extractive and abstractive objectives~\cite{Gehrmann2018BottomUpAS}. While there have been numerous efforts in combining such objectives in traditional RNN networks~\cite{See2017GetTT, Chen2019MultiTaskLF, Lebanoff2018AdaptingTN}, a few studies have explored their efficacy in Transformer-based networks. For instance, \citet{Liu2019TextSW} proposed \textsc{BertSumExtAbs} to utilize extractive objectives from \textsc{BertSumExt} model and further incorporate it with the Transformers decoder to perform abstractive summarization. More recently, \citet{Akiyama2021HieBARTDS} proposed \textsc{Hie-Bart} that adds a self-attention layer for incorporating sentence importance in the encoder. While they augment the encoder with sentential information in the encoder side, the incorporation of such module in the decoder has not been explored in literature. To the best of our knowledge, we are the first to explore this direction on Transformers-based networks in our paper. 

\noindent \textbf{Curriculum Learning.}
Curriculum Learning (CL)~\cite{Bengio2009CurriculumL} has gained growing interest from the research communities during the last decade~\cite{Tay2019SimpleAE, MacAvaney2020TrainingCF, Xu2020CurriculumLF}, although its teaching approach (i.e., learning from easy instances to more difficult ones) known as \textit{incremental learning} dates back to nineties~\cite{Elman1993LearningAD}. The underlying idea of this technique is to provide a training strategy that flows the learning process from easy samples to the harder ones, which results in improving model performance, decreasing training time, and enhancing the model's generalization ability~\cite{Chang2021DoesTO}.  \citet{Bengio2015ScheduledSF} were the first to apply this strategy in the context of sequence prediction using RNN networks through their \textit{scheduled sampling} approach, which gently changes the training process from ground truth tokens to model generated ones during decoding. \citet{Platanios2019CompetencebasedCL} proposed a CL-based NMT framework that decides to visit training samples based on their difficulty and competence state of the model. Sample's \textit{difficulty} is a key concept in this scheme as it is used to distinguish easy examples from the difficult ones. Researchers have used many textual features as the ``difficulty measure'' including n-gram frequency~\cite{Haffari2009ActiveLF}, word rarity and sentence length~\cite{Platanios2019CompetencebasedCL}. Recent works~\cite{Saxena2019DataPA, Cachola2020TLDRES} have made use of confidence-aware approaches that learn the difficulty of training samples and dynamically reweight samples in training process.

\section{Our Approach}
In this section, we  describe the details of our proposed models, including (1) extension of \bart{} model in which a cross-attention layer is added into \bart{} decoder; and (2) our curriculum learning architecture added on the \bart 's Transformer-based framework which upweights easier training samples; hence, increasing their contribution in learning stage. Both of these extensions can be added to the \bart's Transformers network and trained either in joint or independently. 

\begin{figure}
    \centering
    \includegraphics[scale=0.27]{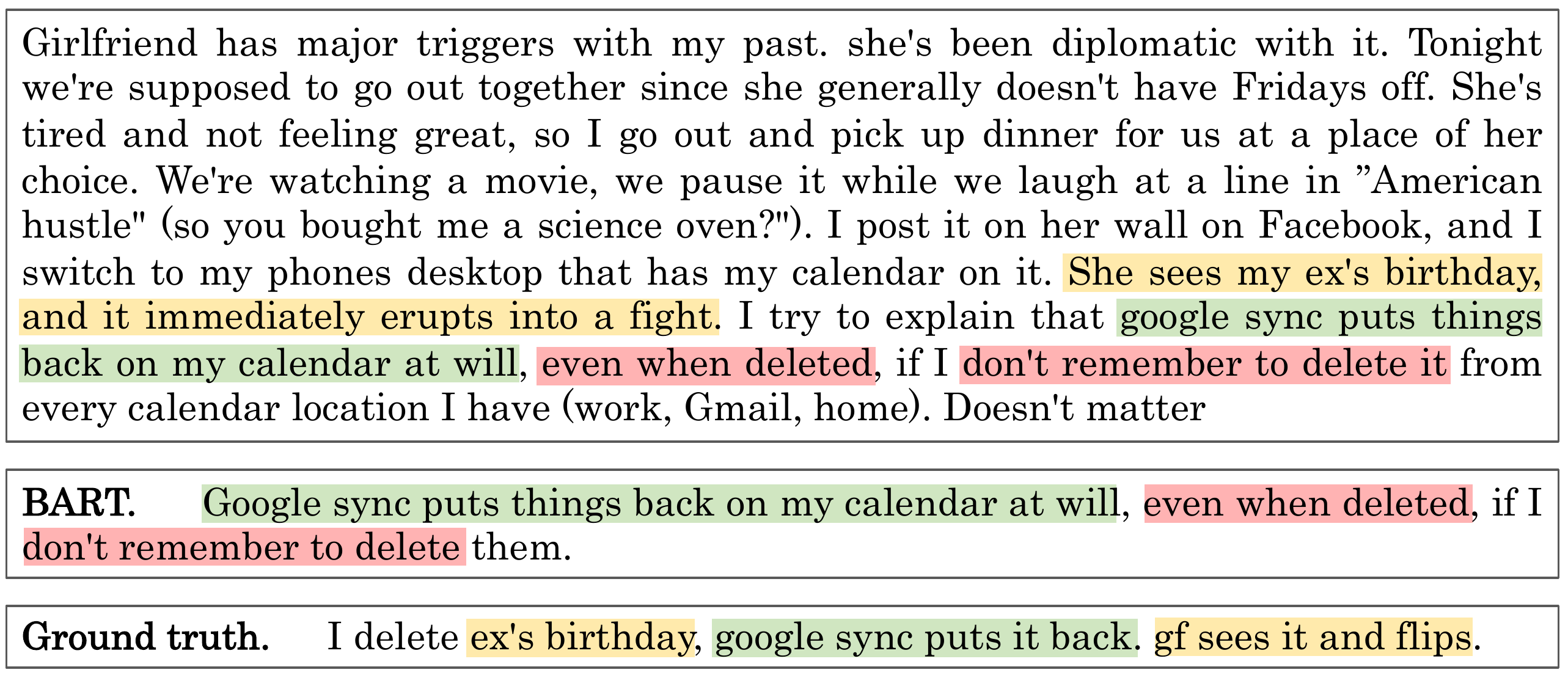}
    \caption{\bart's shortcoming in content selection. Yellow: picked by the ground truth, but skipped by \bart{}, Green: picked by both \bart{} and ground truth, and Red: picked by \bart, but skipped by ground truth}
    \label{fig:bart_shortcoming}
    \vspace{-0.3cm}
\end{figure}

\subsection{Sentence-guided  \textsc{Bart}}
 While \bart{} has been shown to be promising in producing abstractive summaries in virtue of its powerful pre-trained encoder and decoder, it suffers from a major pitfall that restricts model efficacy in content selection. Figure \ref{fig:bart_shortcoming} shows a Reddit post along with \bart's generated and ground truth \tldr, demonstrating such a shortcoming. As observed, while the generated summary appears to be well-written and fluent, it ignores salient source regions and focuses on less important parts of the source. Considering the effectiveness of combining extractive and abstractive objectives, we  extend \bart{} by adding a cross-attention layer to induce sentences' importance at decoding time. To this end, we first define a sequence labelling task, where the goal is to predict sentences' \textit{relative importance} score (i.e., $\mathbf{y}$) such that \bart{} is fine-tuned to learn sentential saliency.  The \textit{relative importance} score is casted as the normalized mean of \rgt{} and \rgl{} scores of source sentences with respect to the \tldr{} summary:
\begin{equation}
     \mathbf{y} =    \text{relative importance($s_i$)} = \frac{ \text{RG}_{\text{2+L}}(s_i)}{\sum\limits_{s_i \in R}{\text{RG}_{\text{2+L}}(s_i)}}
     \label{eq:importance_score}
\end{equation}
where $s_i$ is the sentence in $i$th position, $R$ is the set of post's sentences, and $\text{RG}_{\text{2+L}}(.)$ is a function that takes in a source sentence and outputs the mean of its \rgt{} and \rgl{} scores. For adapting sequence classification task to the sequential sentence tagging problem, we insert \texttt{</s>} tokens (i.e., \texttt{EOS} in \bart{} vocabulary) to the end of each input sentence and then feed it into \bart{} network, similar to \citet{Liu2019TextSW} for \textsc{BertSum}. As \bart{} encodes each input token through its network, the encodings associated with \texttt{</s>} tokens, specifically,  represent input sentences' features preceding them. This is due to the fact that \bart{} uses  \texttt{</s>} tokens' representations as the classification head ~\cite{Lewis2020BARTDS}. After obtaining representations associated with \texttt{</s>} tokens, we process them through a linear layer with Sigmoid classifier to output probabilities as the sentences' importance scores. Formally, let $\mathbf{P}$ be a Reddit post containing sentences $\mathbf{P}=[sent_1, sent_2, ..., sent_i, ..., sent_n]$ and $sent_i=[x_{i1}, x_{i2}, ..., x_{ij}, ..., x_{im}]$.
We frame input $\mathbf{P}$ by adding \texttt{</s>} tokens to the end, as well as \texttt{<s>} to the start of each sentence. In this sense, the modified input to \bart{} network is $\mathbf{P^\prime} = [\texttt{\textbf{<s>}} sent_1 \footnotesize \texttt{\textbf{</s><s>}} sent_2 ... \footnotesize \texttt{\textbf{</s><s>}} sent_n \footnotesize \texttt{\textbf{</s>}}]$ which is then processed through the \bart{} network. The network is trained to predict the importance score (i.e., $\mathbf{y}$ in Eq \ref{eq:importance_score}). By training such a sequence tagger network, we aim to inject an inductive bias to the \bart{} encoder and decoder to get wised up about the source sentences' importance, which will come in handy when generating abstractive summaries. 

\begin{figure}[t]
    \centering
    \includegraphics[scale=0.15]{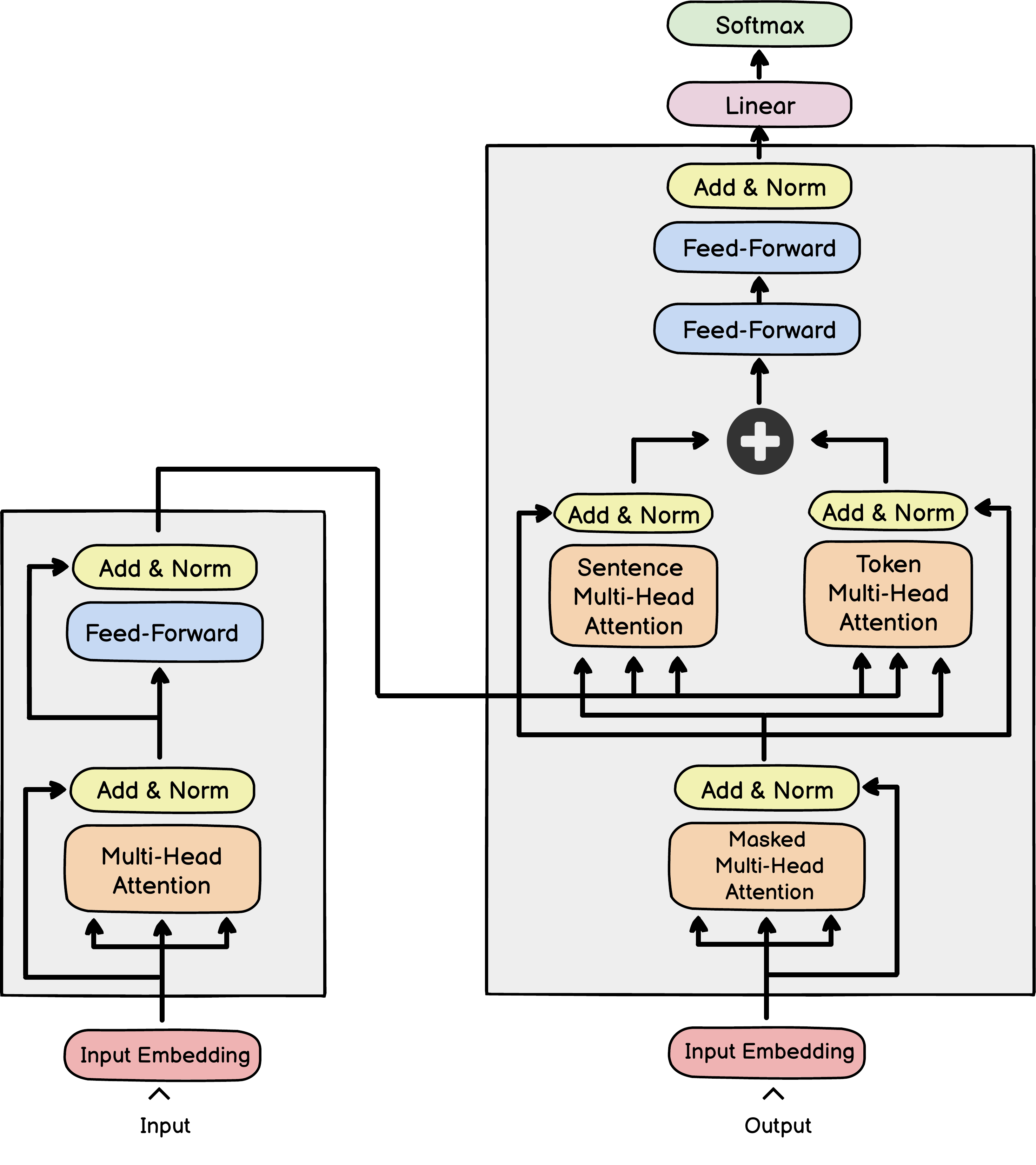}
    \caption{Overview of our summarization model}
    \label{fig:our_model}
    \vspace{-0.7em}
\end{figure}

At the next stage, we design our framework as demonstrated in Figure \ref{fig:our_model} by extending the decoder with an additional cross-attention layer (i.e., Sentence Multi-Head Attention inside the decoder). We use a two-stage fine-tuning approach to this end, where firstly, we fine-tune the encoder module, as well as the Sentence Multi-head Attention on sequence tagging problem. Second, we further fine-tune it on abstractive summarization task. We separate the optimizers of the pre-trained part (i.e., the encoder and Sentence Multi-Head Attention modules) and the decoder. That is, the pre-trained part should be fine-tuned with a lower learning rate, so it is trained with more accurate gradients as the decoder becomes stable. We name this model as \sentsum{} in our experiments.

\subsection{Curricular Learner for \bart{}}
Curriculum learning (CL)~\cite{Bengio2009CurriculumL} is a training paradigm to improve the performance and generalization of the learner models based on the idea that easy samples should be visited before the difficult ones during the training~\cite{Castells2020SuperLossAG}. This is due to the fact that when the model starts off with easier samples on early stages of training, the risk of getting stuck in local optima is reduced as most loss functions in deep neural networks are highly non-convex~\cite{Chang2021DoesTO}, and hard to converge. Considering the applicability of curriculum learning in training large-scale networks, we aim to use it in our summarization framework. Before incorporating the curriculum learning strategy into our model's training stage, we first need to define the \textit{difficulty} metric to distinguish the hardness of samples.

In practice, estimating a prior difficulty for each sample is considered a complex task, so we propose to discriminate the samples with progressive signals ---such as the respective sample loss at each training iteration--- in training process. In this context, CL is achieved by predicting the difficulty of each sample at the training iterations in the form of a weight, such that difficult samples receive lower weights during the early stages of training and vice versa. To model the curriculum, we propose to use \superl~\cite{Castells2020SuperLossAG} which is a generic loss criterion built upon the task loss function.
More specifically, \superl{} is a task-agnostic confidence-aware loss function that takes in two parameters: (1) the task loss $\losslrg_i =  \loss(y_i, \widehat{y_i})$, where $y_i$ is neural network's (i.e., \bart's generated summary) output and $\widehat{y_i}$ is the gold label (i.e., ground-truth summary); and (2) $\sigma_i$ as the confidence parameter of the $i$th sample. \superl{} is framed as $\text{L}_{\lambda}(\losslrg_i, \sigma_i)$ and computed as follows,

\begin{equation}
    \text{L}_{\lambda}(\losslrg_i, \sigma_i)= (\losslrg_i - 	\tau) \sigma_i + \lambda (\log \sigma_i)^2
\end{equation}
in which $\lambda$ is the regularization parameter, and $\tau$ is the running or static average of task loss (i.e., $\losslrg$) during the training. While \superl{} provides a well-defined approach to curriculum learning strategy, learning $\sigma$ parameter is not tractable for tasks with abundant training instances such as text summarization. To circumvent this issue and hinder imposing new learnable parameters, \superl{} suggests using the converged value of $\sigma_i$ at the limit,

\begin{align}
\label{eqn:eqlabel}
\begin{split}
\sigma^{*}_{\lambda} (\loss_i) &= \argmin_{\sigma_i} \text{L}_{\lambda} (\loss_i, \sigma_i)
\\
 \text{SL}_\lambda(\loss_i) &= \text{L}_\lambda(\loss_i, \sigma^{*}_{\lambda}(\loss_i, \sigma_i)) = \min_{\sigma_i} \text{L}_\lambda (\loss_i, \sigma_i),
\end{split}
\end{align}

Using this technique, the confidence parameters are not required to be learned during the training. \citet{Castells2020SuperLossAG} found out that $\sigma^{*}_{\lambda} (\loss_i)$ has a closed-form solution, computed as follows,

\begin{equation}
     \sigma^{*}_{\lambda} (\loss_i) = e ^ {-W (\frac{1}{2} \max (-\frac{2}{e}, \beta))}, \beta = \frac{\loss_i - \tau}{\lambda}
\end{equation}
in which $W$ is the Lambert W function. With this in mind, \superl{} upweights easier samples dynamically during the training; hence, providing a curriculum learning approach to summarization. We call this model \currsum{}. We also experiment with a combination of \sentsum{} and \currsum{} models and name it \currsentsum{} throughout our experiments.

\section{Experimental Setup}
\label{sec:res}

\subsection{Datasets}
We use two Reddit summarization datasets including Reddit TIFU dataset~\cite{Kim2019AbstractiveSO} and Webis-TLDR-17~\cite{volske-etal-2017-tl}, as well as two well-known news summarization datasets including CNN/DM~\cite{See2017GetTT}, and XSum~\cite{Narayan2018DontGM} throughout our experiments. Reddit datasets are gathered from Reddit discussion forums containing 42k (Reddit TIFU) and 4M (Webis-TLDR-17) instances with source (i.e., post's text) and a \tldr{} summary written by human. We use 80\% (33,705)-10\% (4,214)-10\% (4,214), and 99\% (3,771,432)-1\% (38,483)-1\% (38,484) random train-val-test splits for Reddit TIFU and Webis-TLDR-17, respectively. For the news summarization datasets, we use the  splits suggested by their original papers.

\subsection{Comparison}
We compare our model against various extractive and abstractive state-of-the-art baselines. The description of each baseline is outlined below. 


\noindent \textsc{\textbf{BertSumExt}}~\cite{Liu2019TextSW}: the extractive variant of \textsc{BertSum} that inserts external \texttt{[CLS]} tokens between sentences. The representations are further used with a Sigmoid classifier to compute sentence extraction probabilities.

\noindent \textsc{\textbf{BertSumAbs}}~\cite{Liu2019TextSW}: the abstractive variant of \textsc{BertSum} that uses a Transformer-based encoder-decoder architecture. The encoder is \textsc{Bert}, but decoder is pre-trained from scratch. 

\noindent \textsc{\textbf{BertSumExtAbs}}~\cite{Liu2019TextSW}: a two-stage fine-tuning approach, for which \textsc{BertSumExtAbs} first fine-tunes encoder on extractive summarization, then it is fine-tuned along with decoder for abstractive summarization task. 

\noindent \textsc{\textbf{MatchSum}}~\cite{zhong-etal-2020-extractive}: an extractive framework which matches source text with candidate summaries in a semantic space. Unlike commonly used extractive summarizers, \textsc{MatchSum} does not rely on extracting sentences individually, instead it selects a set of sentences (i.e., candidate summary) that has the maximum semantic similarity with the ground-truth summary.

\noindent \textsc{\textbf{Pegasus}}~\cite{zhang2019pegasus}: an abstractive  model that defines a new pre-training task as Gap Sentence Generation (GSG), in which key source sentences are masked out, and the network learns to generate the missing sentences.

\noindent \textsc{\textbf{Bart}}~\cite{Lewis2020BARTDS}: an abstractive model that  uses a pre-trained encoder-decoder architecture, unlike \textsc{Bert} that only utilizes a pre-trained encoder.

\noindent \textsc{\textbf{NeuTopicSumm}}~\cite{Nguyen2021EnrichingAC}: an abstractive baseline that incorporates neural topic model into summarization framework.

\noindent \textsc{\textbf{BART+R3F}}~\cite{Aghajanyan2021BetterFB}: a summarization method that proposes a new fine-tuning technique; R3F method replaces previously proposed adversarial objectives with parametric noise, and discourages representation change during fine-tuning without degrading the performance. 

\noindent \textsc{\textbf{BART+MUPPET}}~\cite{Aghajanyan2021MuppetMM}: an architecture that introduces an additional large-scale learning stage (i.e., pre-finetuning) between the pre-training and fine-tuning stages, using a large-scale collection of datasets proposed on different tasks. \textsc{MUPPET} is designed to encourage representation learning for increasing the generalization of language models.

\subsection{Implementation details}
We extend the Huggingface's Transformer's library ~\footnote{\url{https://github.com/huggingface/transformers}}~\cite{Wolf2020Transformers}  to implement our models, and make the code publicly available to expedite future research~\footnote{\textit{HTTP}}. We train all of our models for 8 epochs (Reddit TIFU) and 5 epochs (Webis-TLDR-17, CNN/DM, and XSum) and use the checkpoint that achieves the best \textsc{Rouge-L} score in the validation for the inference. 
AdamW optimizer~\cite{Loshchilov2019DecoupledWD} initialized with learning rate of $3e-5$, $(\beta_1, \beta_2)= (0.9, 0.98)$, and weight decay of 0.01 is used for all of our summarization models, as well as for \textsc{Bart}. Specifically, to train \sentsum{}, we use a lower learning rate of $1e-5$ for the pre-trained part. Cross-entropy loss is used for all models, except for pre-training \sentsum{} where we use the Mean Squared Error (MSE) loss function. For \textsc{BertSum} variants, we use the main code-base~\footnote{\url{https://github.com/nlpyang/PreSumm}} and default hyper-parameters suggested by the original paper~\cite{Liu2019TextSW}. To keeping track of the learning process, we use Weights \& Biases~\cite{wandb} toolkit. 

\section{Results}
\textbf{Automatic Evaluation.}
Table \ref{tab:final} reports the performance of the baseline models along with our models' in terms of \textsc{Rouge} score variants~\cite{Lin2004ROUGEAP} over (a) Reddit TIFU dataset. As indicated, our best model is \currsum{} that uses \superl{} directly on top of the \bart{} model and is a clear improvement over most of the baselines across all metrics. 
Specifically, \currsum{} outperforms its ground baseline that has no curriculum (i.e., \bart{}) by relative improvements of 5.2\%, 11.22\%, 5.4\% for \rgo, \rgt, \rgl, respectively, 
on Reddit TIFU. While \currsum{} achieves competitive performance in terms of \rgo{} with \textsc{BART+R3F} and \textsc{BART+MUPPET}, it lags behind them on \rgt{} and \rgl. These differences may be explained by the fact that \textsc{BART+R3F} is computationally more intensive than our curriculum-based model~\footnote{BART+R3F adds an additional forward pass (FP) to compute symmetric KL divergence term for measuring parametric noise.}, and \textsc{BART+MUPPET} performs the best presumably due to its additional large-scale learning stage (i.e., pre-finetuning) that is not considered in \bart{} and \currsum~\footnote{BART+MUPPET is not released at the time of writing this study; hence, we could not report its performance when curriculum is applied.}. Interestingly, while augmenting the decoder with sentence cross-attention module (i.e., \sentsum) is marginally better than the \bart{} baseline, training it with curriculum learning \superl{} function (i.e., \currsentsum) further improves the performance. This finding provides a compelling evidence for the usefulness of curricular training strategy applied in summarization task. Comparing abstractive vs. extractive models, we observe a noticeable performance gap denoting that Reddit TIFU includes rather abstractive summaries than extractive.

\begin{table}[t]
\centering
\scalebox{0.8}{
   \begin{tabular}{l@{\hspace{1cm}}lll}
       \toprule
     Model                    & \textsc{RG-1}  & \textsc{RG-2}  & \textsc{RG-L}  \\
    \midrule
     \textsc{BertSumExt~\citeyearpar{Liu2019TextSW}}             &  20.32 & 4.81 & 13.77 \\
     \textsc{BertSumAbs}~\citeyearpar{Liu2019TextSW}                             & 21.92 & 4.99 & 14.21 \\
     \textsc{BertSumExtAbs}~\citeyearpar{Liu2019TextSW}                 & 22.14 & 6.01 & 14.66 \\
      \textsc{MatchSum}~\citeyearpar{zhong-etal-2020-extractive}    & 25.09     & 6.17     & 20.13     \\

     \textsc{Pegasus}~\citeyearpar{zhang2019pegasus}    & 26.63     & 9.01     & 21.60     \\
     
     \textsc{Bart}~\citeyearpar{Lewis2020BARTDS} & 28.80 & 9.02 & 23.02  \\
     
     \textsc{NeuTopicSumm}~\citeyearpar{Nguyen2021EnrichingAC}  &  27.96 &  9.43 & 23.08 \\
     
     \textsc{BART+R3F}~\citeyearpar{Aghajanyan2021BetterFB}  &  \textbf{30.31} &  {10.98} & {24.74}  \\
    
    \textsc{BART+MUPPET}~\citeyearpar{Aghajanyan2021MuppetMM}  &  \textbf{30.30} &  \textbf{11.25} & \textbf{24.94} \\
     
    \noalign{\vskip 2.5pt}
\hdashline[1pt/2pt]
\noalign{\vskip 2.5pt}

    \sentsum{} (Ours) & 29.09 & 9.14& 23.39\\
    \currsum{} (Ours) & \textbf{30.32}& {10.16}& {24.27} \\
    \currsentsum{} (Ours) & {29.57} & {9.81}& {23.61}\\
    \bottomrule  
\end{tabular}

}

\caption{\textsc{Rouge} results on test set of Reddit TIFU dataset. We bold the best numbers and the numbers within 0.15 of the best.}
\label{tab:final}
\end{table}




\begin{figure}[t]

    \begin{tabular}{ll}
    \hspace{-0.7cm}
    \includegraphics[scale=0.205]{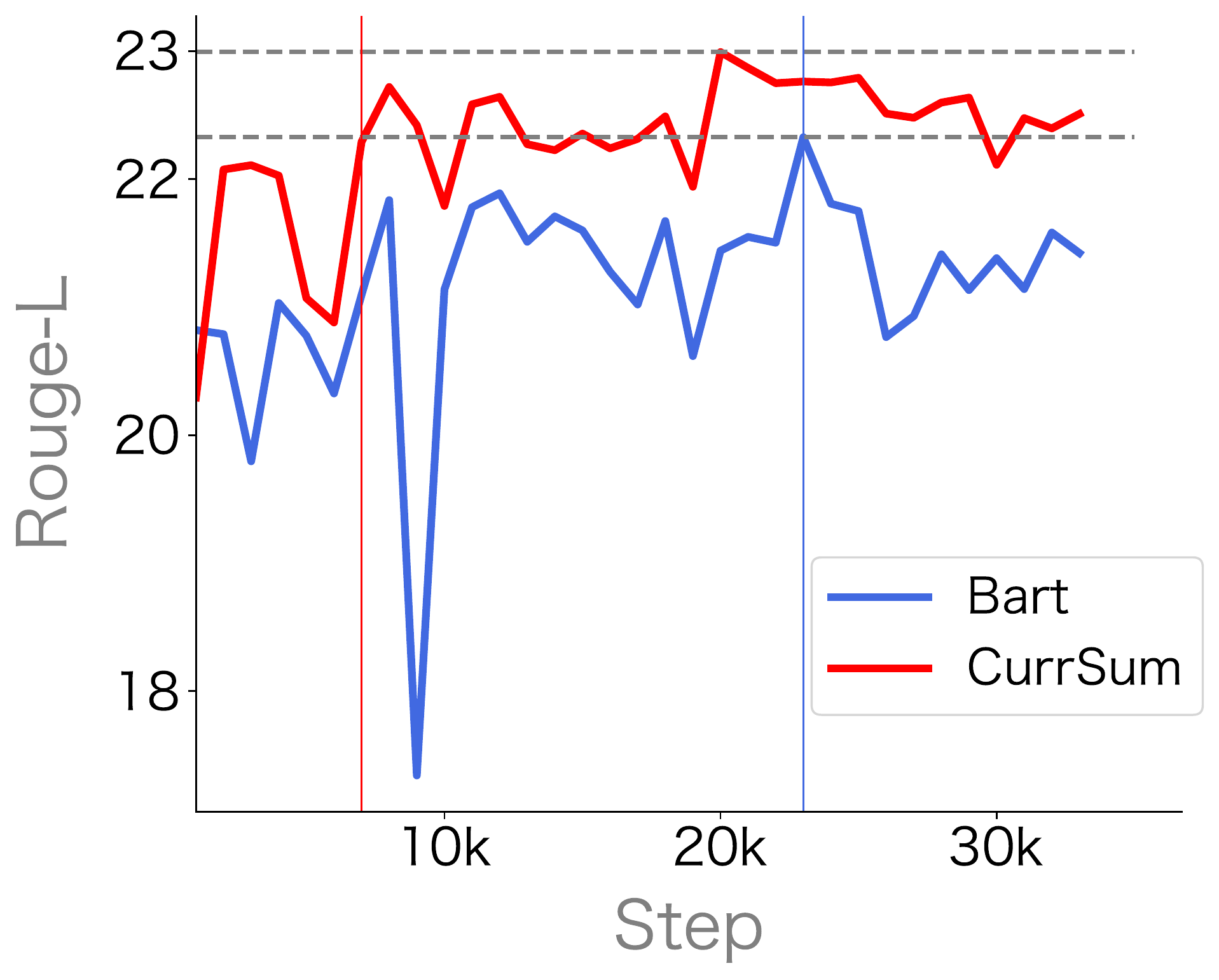}
    
    &
    \hspace{-0.5cm}
    \includegraphics[scale=0.205]{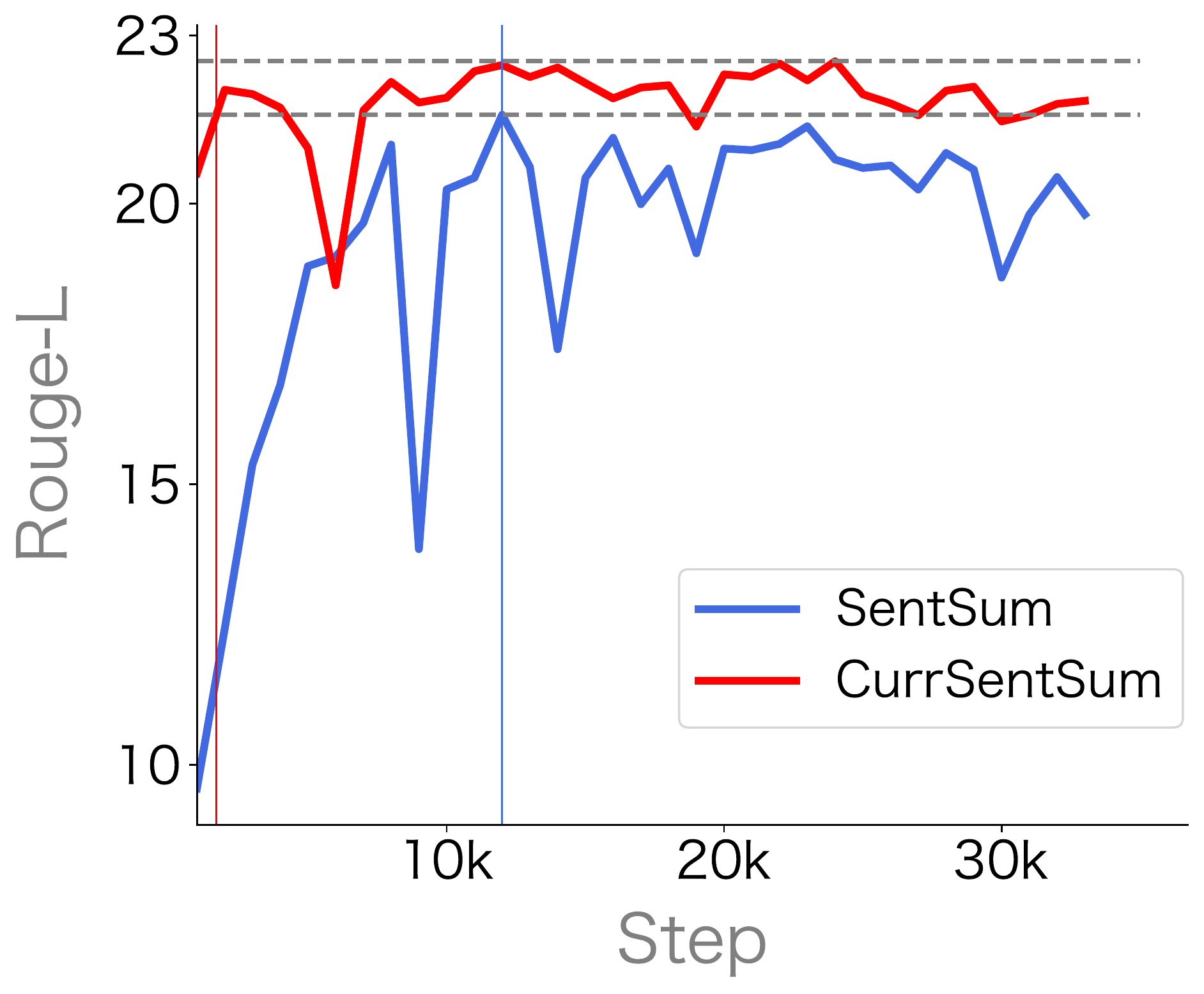}

    \end{tabular}

    \caption{Plots demonstrating the validation performance of various models with increasing steps of training of Reddit TIFU dataset. Blue lines show the performance when no curriculum is used. Red lines represent the performance when the curriculum is added. Vertical lines indicate the step where the models achieve the \rgl{} score that the baselines attain at convergence.}
    \label{fig:val_performance}
    \vspace{-0.3cm}
\end{figure}

We further investigate the validation performance of the summarization models in the presence and absence of the curriculum learning strategy. As shown in Figure \ref{fig:val_performance}, models trained with curriculum strategy (i.e., \currsum{} and \currsentsum) tend to converge faster, and perform better in comparison with their respective baselines (i.e., \bart{} and \sentsum) that are without curriculum learning. 
Looking at Figure \ref{fig:val_performance}, the efficiency of curriculum strategy is quite remarkable considering the scores and convergence steps (i.e., vertical red lines) of the curriculum-equipped models.


\noindent\textbf{What is the effect of curriculum learning on cross-domain datasets? } To measure the cross-domain performance of models supplied with curriculum learning strategy, we compare \bart's performance (as a transformer-based  state-of-the-art summarizer) vs. a variant of \bart{} with curriculum strategy (i.e., our \currsum{} model) on three summarization datasets: Webis-TLDR-17, CNN/DM, and XSum. Results, summarized in Table \ref{tab:tab2}, are a strong indication of effectiveness of curriculum learning on datasets from various domains (i.e., Social Media and News) as we observe a consistent improvement compared to the \bart{} baseline.

\begin{figure*}[t]
\centering
\bgroup
\setlength{\tabcolsep}{0pt}
\renewcommand{\arraystretch}{0}
\begin{tabular}{cccccc}
\includegraphics[scale=0.14]{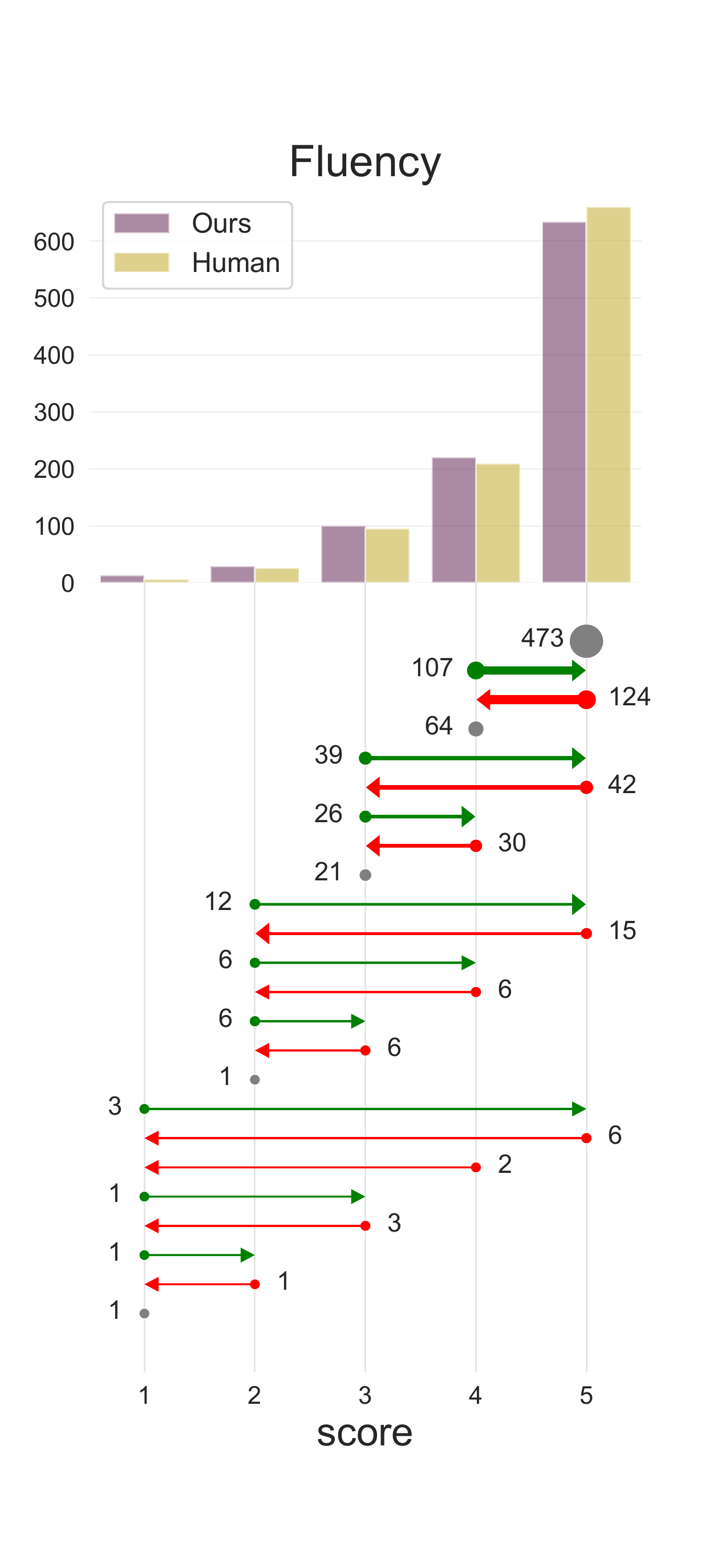} &
\includegraphics[scale=0.14]{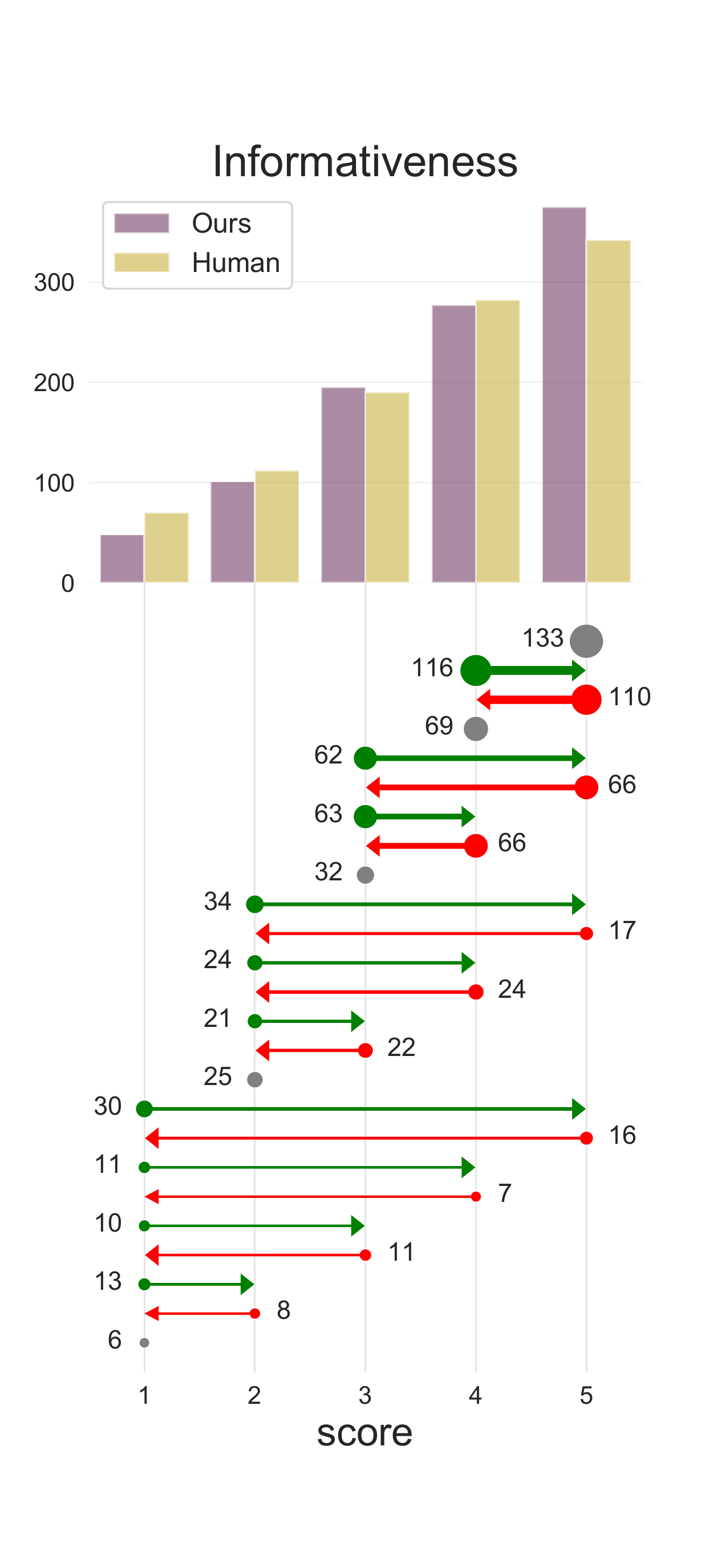} &
\includegraphics[scale=0.14]{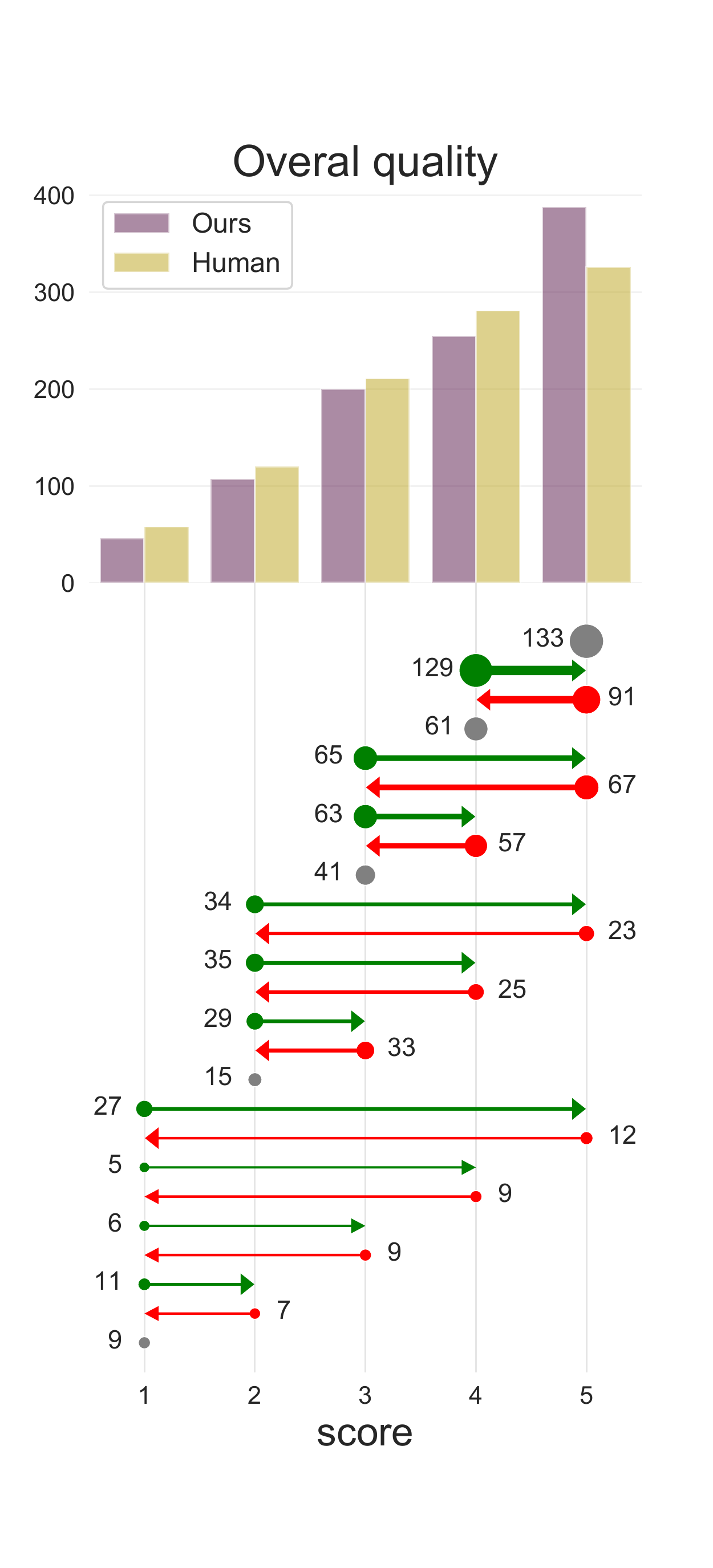} &
\includegraphics[scale=0.14]{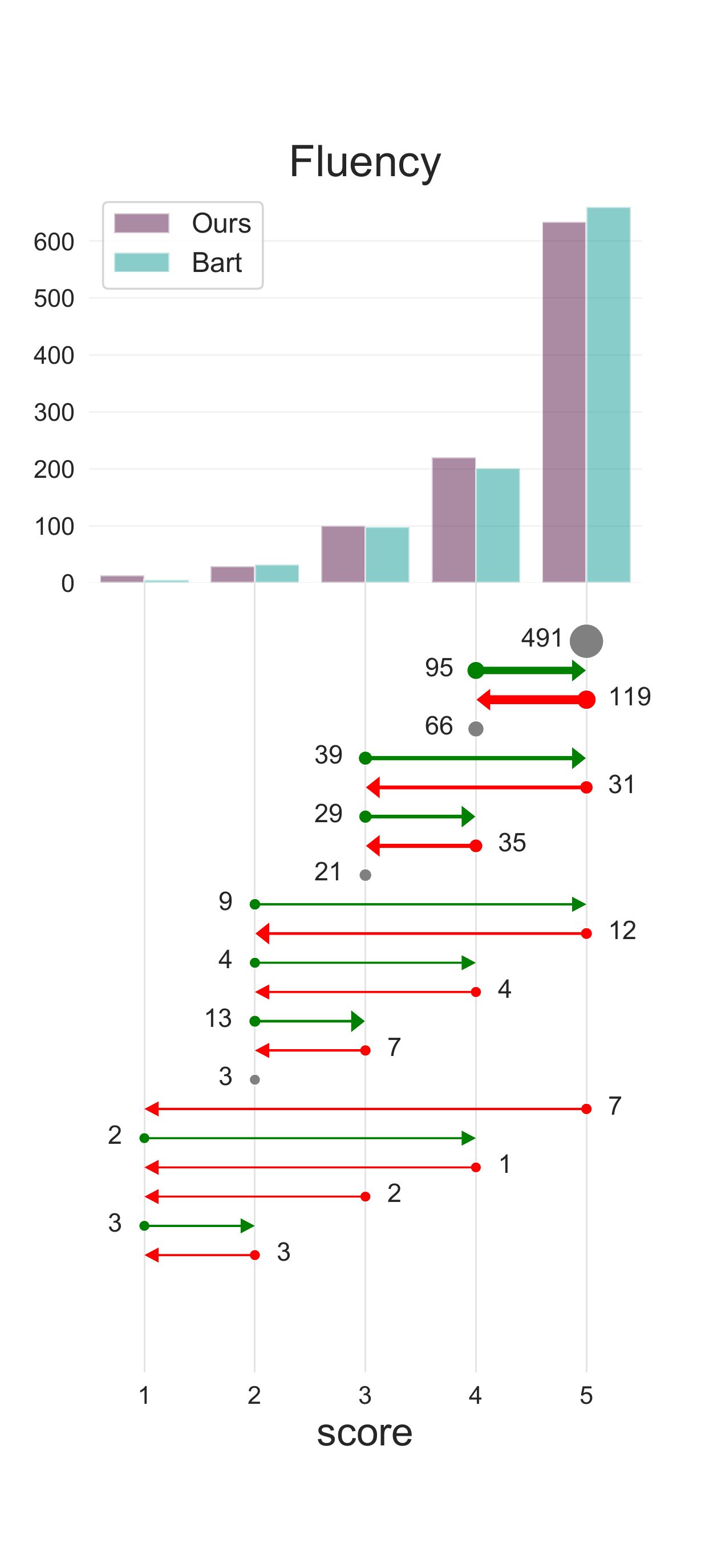} &
\includegraphics[scale=0.14]{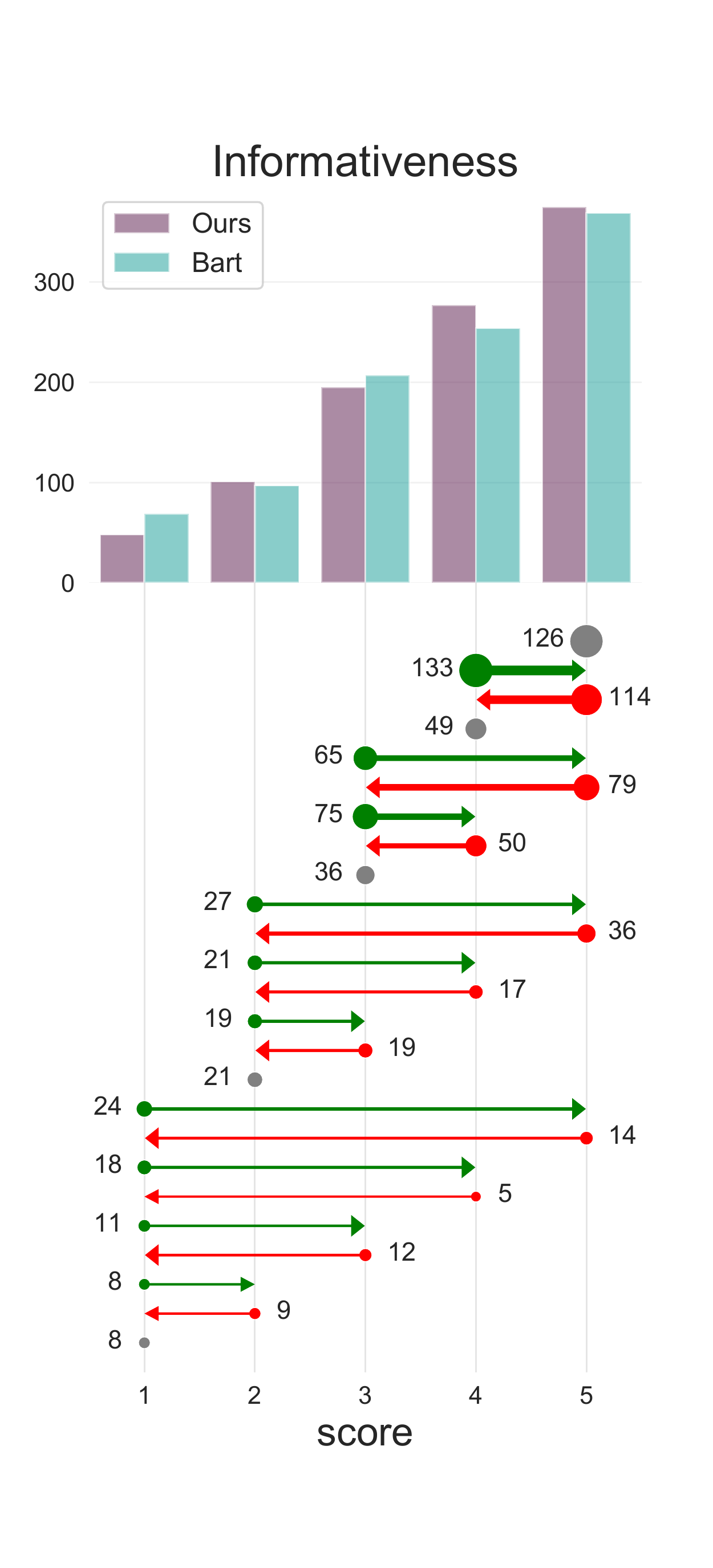} &
\includegraphics[scale=0.14]{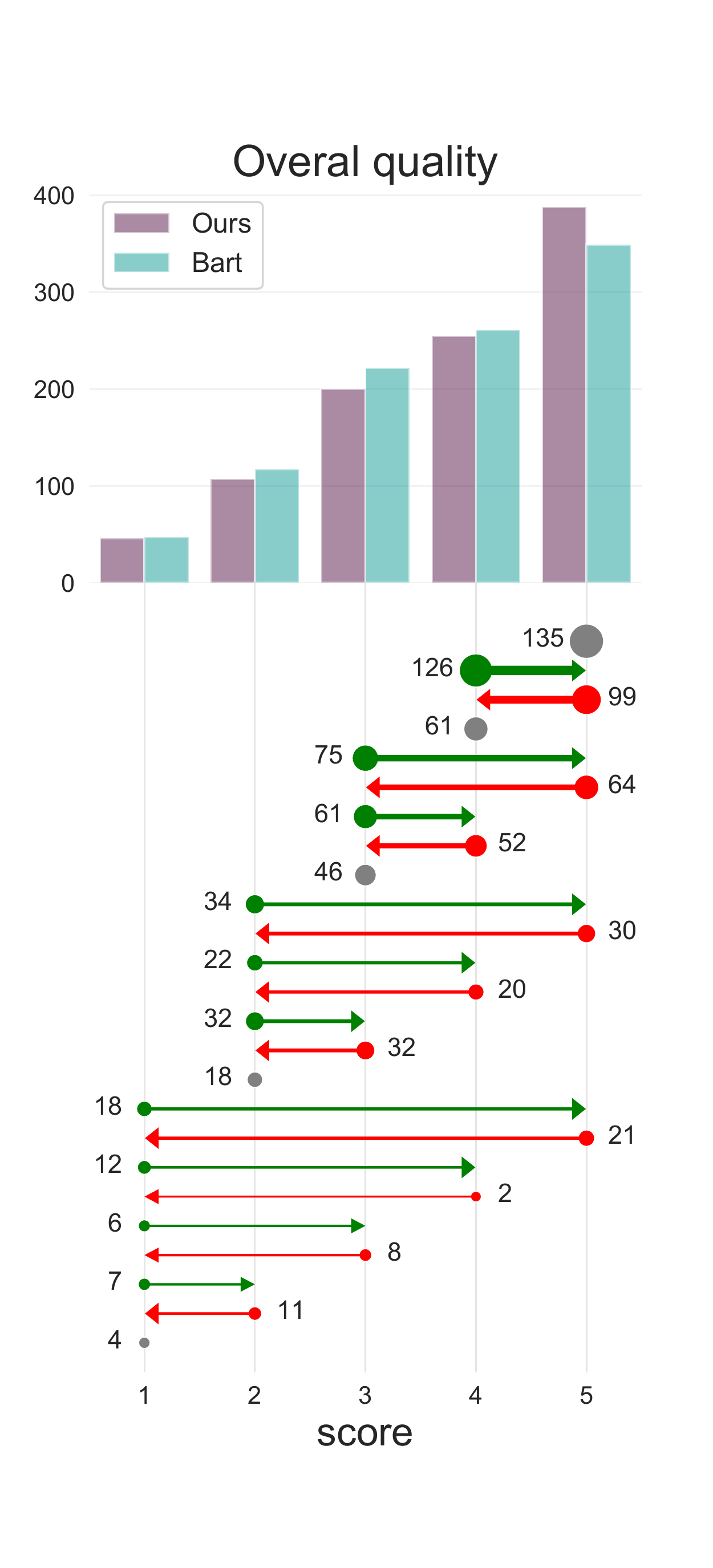} \\
(a) & (b) & (c) & (d) & (e) & (f) \\
\end{tabular}
\vspace{-0.5em}
\caption{Histograms and arrow plots depicting score transition between 200 manually-scored \tldr{} summaries. (a--c) and (d--f) show our system's comparison with human and \bart{} baseline system, respectively. Although comparable in terms of fluency metric with the baseline and human parity, our model makes a strong gain to improve summary's informativeness and overall quality.}
\label{fig:arrows}
\egroup
\end{figure*}

\begin{table}[t]
    \centering
\scalebox{0.8}{
\begin{tabular}{c}
    
   \begin{tabular}{l@{\hspace{1.2cm}}lll}
    \toprule
     Model                    & \textsc{RG-1}  & \textsc{RG-2}  & \textsc{RG-L}  \\
         \midrule
    \textsc{Bart}~\cite{Lewis2020BARTDS}  &  20.73 &  6.79 & 17.22 \\
     
    \currsum{} (Ours)  & \textbf{22.12}& \textbf{6.91} &  \textbf{17.97} \\
    
    \bottomrule 
    \end{tabular}
    
        \vspace{0.1cm} \\
         (a) Webis-TLDR-17 \vspace{0.2cm} \\
          
  \begin{tabular}{l@{\hspace{1.2cm}}lll}
    \toprule
     Model                    & \textsc{RG-1}  & \textsc{RG-2}  & \textsc{RG-L}  \\
         \midrule
    \textsc{Bart}~\cite{Lewis2020BARTDS}  &  44.16 &  21.28 & 40.90 \\
     
    \currsum{} (Ours)  & \textbf{44.49}& \textbf{21.53} &  \textbf{41.19} \\
    
    
    \bottomrule 
    
    \end{tabular}
    
    \vspace{0.1cm} \\
         (b) CNN/DM \vspace{0.2cm} \\
    
    \begin{tabular}{l@{\hspace{1.2cm}}lll}
    \toprule
     Model                    & \textsc{RG-1}  & \textsc{RG-2}  & \textsc{RG-L}  \\
         \midrule
    \textsc{Bart}~\cite{Lewis2020BARTDS} &  45.14 &  22.27 & 37.25 \\
     
    \currsum{} (Ours)  & \textbf{45.41}& \textbf{22.39} &  \textbf{37.41} \\
    
    \bottomrule  
    
    \end{tabular}
       \vspace{0.1cm} \\
         (c) XSum  
    \end{tabular}
    
}

    \caption{Results when applying curriculum learning for \bart{} on cross-domain datasets. }
    \label{tab:tab2}
\vspace{-0.4cm}
\end{table}

\noindent \textbf{Human evaluation.}
A few studies have recognized the limitation of widely adopted \textsc{Rouge} metric as it is biased toward surface lexical similarities~\cite{Ng2015BetterSE, Cohan2016RevisitingSE}. To get insights into the qualities of our proposed models, we performed a human evaluation over a random set of system-generated summaries. To this end, we randomly sampled 200 cases from Reddit TIFU's test set, each consisting of the post's source text along with blinded \tldr's from (1) author-written references; (2)  \bart{} baseline; and (3) our \currsentsum{} model. The choice of \currsentsum{} lies in the fact that we want to evaluate the effect of both sentence attention and curriculum learning in our human evaluation process. To prevent potential bias, we randomly shuffled the ordering of \tldr s provided to evaluators. Following prior work~\cite{Grusky2018NewsroomAD, Zhang2020OptimizingTF, Cho2021StreamHoverLT}, we define three metrics: (1) \textbf{Fluency: } is the \tldr{} well-written and easy to understand? (2) \textbf{Informativeness: } does the \tldr{} provide useful information about source? (3) \textbf{Overall quality: } overall, is the \tldr{} of good quality in terms of content (both fluency and informativeness), and correctness?  We then had five human evaluators score each of the provided examples on a scale of [1-5] (worst to best) in terms of the criteria mentioned above. The evaluators were familiar with data science and annotation and were hired through the Upwork~\footnote{\url{https://www.upwork.com}} freelancing platform.

\begin{table}[t]
    \centering
    \begin{tabular}{lccc}
        \toprule
         System & Fluency & Info. & Overall  \\
         \midrule
         \bart{} & 4.48 & 3.76 & 3.75   \\
         \currsentsum & 4.45 & \textbf{3.91} & \textbf{3.98}
          \\ 
          Human & \textbf{4.50} & 3.72  &  3.70 \\ 
         \bottomrule
    \end{tabular}
    \caption{Results of the human evaluation comparing three systems in terms of {Fluency}, {Informativeness}, and {Overall quality}. Winning scores are shown in bold.}
    \label{tab:human_evaluation}
\end{table}

Table \ref{tab:human_evaluation} shows the average score gained by each system in terms of the aforementioned qualitative criteria. Comparing our model against other systems, we find that: (1) in terms of fluency, our model and \bart{} are quite comparable, with human summaries being the best; (2) we calculated the average \tldr{} length (in tokens) of the test set and obtained 22.9 (Human), 23.6 (\bart), and 19.8 (\currsentsum). Interestingly, despite generating shorter summaries, our model outperforms both \bart{} and human-written \tldr s on informativeness which shows that it efficiently selects the useful information from the original text, generating them in a comparably shorter text. Looking into the outperformed cases on informativeness, we noticed that the annotators tend to give a higher score to a summary if it provides the most important information within a concise text (i.e., providing only to the point information akin to the definition of \tldr), which is generally achieved by our model.
(3) our model is also more preferable in terms of overall quality compared to \bart{} and human summaries, with relatively large gap. Overall, it is interesting that summarization models are becoming comparable (sometimes even more preferable) to Human \tldr s on informativeness and overall quality, substantiating the recent success of pre-trained language models as also shown by~\citet{Fabbri2021SummEvalRS}. We further computed the Fleiss' Kappa~\cite{Fleiss1971MeasuringNS} inter-rater agreement for the qualitative metrics, and obtained 10\%, 27\%, 22\% correlation scores for fluency, informativeness, and overall quality, respectively. These correlations are considered to be ``slight'' for fluency and ``fair'' for informativeness, and overall quality with regard to the Fleiss' range interpretation~\cite{Landis1977TheMO}. Table~\ref{tab:agreement} shows the system-wise Fleiss' agreement over three metrics, expressing that agreement rates on our system summaries are stronger than the others.

\begin{table}[t]
    \centering
        \scalebox{0.85}{
    \begin{tabular}{lccc}
         \toprule
         System & Fluency & Info. & Overall  \\
         \midrule
         \bart{} & 12\% & 26\% & 21\% \\
         \currsentsum & \textbf{14\%} & \textbf{33\%} & \textbf{25\%} \\
         Human & 11\% & 24\% & 20\% \\
         \bottomrule
    \end{tabular}
    }
    
    \caption{System-wise Fleiss' kappa agreement}
        \vspace{-1em}

    \label{tab:agreement}
\end{table}

Inspired by ~\citet{MacAvaney2019SIG, Sotudeh2020AttendTM}, we plot histograms and arrow plots for our human evaluation in Figure \ref{fig:arrows}. The histograms show the score distributions gained by each model, and arrow plots demonstrate the score transition (i.e., how scores changed) on the provided samples. The head of each arrow shows our system's score, which makes a transition to its head showing the score gained by the other systems (either human or \bart). The count of the samples that have made a specific score transition is shown next to the arrows' tail.  As indicated, it is observable that our system makes a strong gain in informativeness and overall quality metrics over the other two systems while staying competitive in terms of fluency. The improvement is specifically considerable in enhancing scores from 4 to 5 in informativeness and overall quality metrics.

\noindent \textbf{Qualitative analysis.}
In order to provide insights into qualities of our model vs baselines, we did further evaluation over the annotated samples. We found out that (1) our model performs superior at collecting the key information from the source when there is an overload of important information in the source text, while most gold \tldr s contain just a few sentences (less than 3) that only include the most important information. This behaviour of our model enables it to take advantage of informativeness and overall quality metrics. (2) human-written \tldr s receive relatively a lower score compared to our model's in terms of informativeness and overall quality when the human-written \tldr{} contains entailment/conclusion from the source, although it might not be present in the source text. (3) interestingly, as the system \tldr s become lengthy, the annotators tend to give a lower score in terms of informativeness and overall quality metrics. This might be due to the fact that longer summaries encompass a high proportion of source information regardless of their saliency. 
\section{Conclusion}
 While neural transformer-based summarization models have shown to be promising, they suffer from two shortcomings, namely \textit{content selection}, and \textit{inefficient training process}. In this paper, we explore two approaches to address these issues. Firstly, we propose to tackle the content selection problem by augmenting the decoder via a sentence cross-attention layer such that the decoder becomes aware of sentence saliency. Secondly, we incorporate a confidence-aware curriculum learning approach to the summarization framework in the hope of increasing model's generalization, achieving faster convergence, and ultimately improving model performance. Our automatic evaluations over various data collections from different domains and human evaluations show the effectiveness of our model.

\bibliography{main}
\bibliographystyle{acl_natbib}




\end{document}